\documentclass[letterpaper]{article} 
\usepackage{aaai2026}  
\usepackage{times}  
\usepackage{helvet}  
\usepackage{courier}  
\usepackage[hyphens]{url}  
\usepackage{graphicx} 
\usepackage{natbib}  
\usepackage{caption} 
\usepackage{algorithm}
\usepackage{algorithmic}
\usepackage{subfig}

\usepackage{booktabs}%
\usepackage{amsmath,amssymb,amsfonts}%

\usepackage{newfloat}
\usepackage{listings}
\DeclareCaptionStyle{ruled}{labelfont=normalfont,labelsep=colon,strut=off} 
\floatstyle{ruled}
\newfloat{listing}{tb}{lst}{}
\floatname{listing}{Listing}
\title{ITPP: Learning Disentangled Event Dynamics in Marked Temporal Point Processes}
\author {
    Wang-Tao Zhou\textsuperscript{\rm 1},
    Zhao Kang\textsuperscript{\rm 1},
    Ke Yan\textsuperscript{\rm 1}\thanks{Corresponding author},
    Ling Tian\textsuperscript{\rm 1\rm 2}
}
\affiliations {
    \textsuperscript{\rm 1}University of Electronic Science and Technology of China\\
    \textsuperscript{\rm 2}Shenzhen Institute for Advanced Study, UESTC\\
    wtzhou@std.uestc.edu.cn, zkang@uestc.edu.cn, kyan@uestc.edu.cn, lingtian@uestc.edu.cn
}

\usepackage{bibentry}

\begin{document}

\maketitle

\begin{abstract}
Marked Temporal Point Processes (MTPPs) provide a principled framework for modeling asynchronous event sequences by conditioning on the history of past events. However, most existing MTPP models rely on channel-mixing strategies that encode information from different event types into a single, fixed-size latent representation. This entanglement can obscure type-specific dynamics, leading to performance degradation and increased risk of overfitting.
In this work, we introduce ITPP, a novel channel-independent architecture for MTPP modeling that decouples event type information using an encoder-decoder framework with an ODE-based backbone. Central to ITPP is a type-aware inverted self-attention mechanism, designed to explicitly model inter-channel correlations among heterogeneous event types. This architecture enhances effectiveness and robustness while reducing overfitting.
Comprehensive experiments on multiple real-world and synthetic datasets demonstrate that ITPP consistently outperforms state-of-the-art MTPP models in both predictive accuracy and generalization.
\end{abstract}


\section{Introduction}
\noindent 
The ability to accurately predict future events is of paramount importance across numerous real-world applications,  such as healthcare \cite{healthcare}, finance \cite{finance}, traffic \cite{traffic}, etc. While traditional predictive models often focus on whether an event will occur within a discrete time window, many critical systems operate in continuous time, where the precise timing of an event is as crucial as its occurrence. Modeling such fine-grained temporal event dynamics requires a framework that can capture the stochastic evolution of event times and their associated characteristics. Marked Temporal Point Processes (MTPP) have emerged as a powerful and principled tool for this purpose, providing a mathematical foundation for modeling the joint probability distribution over the time and mark of the next event, conditioned on the history of past events.

MTPP models have garnered considerable research attention in recent years. A substantial portion of these models leverage variants of Recurrent Neural Networks (RNNs) \cite{rmtpp,nhp,lognormmix,ctpp} or self-attention mechanisms \cite{SAHP,THP,attnhp} to encode the sequential patterns within event sequences. More recently, Neural Ordinary Differential Equations (ODEs) have gained traction for modeling temporal irregularities. Following this trend, several ODE-based MTPP models have been proposed \cite{gruode,njsde,nstpp,gstpp}, which focus on designing sophisticated drift and jump networks to simulate the infinitesimal patterns of state evolution over time.

\begin{figure}[ht]

\centering
\subfloat[Channel mixing]{\includegraphics[height=2.1in]{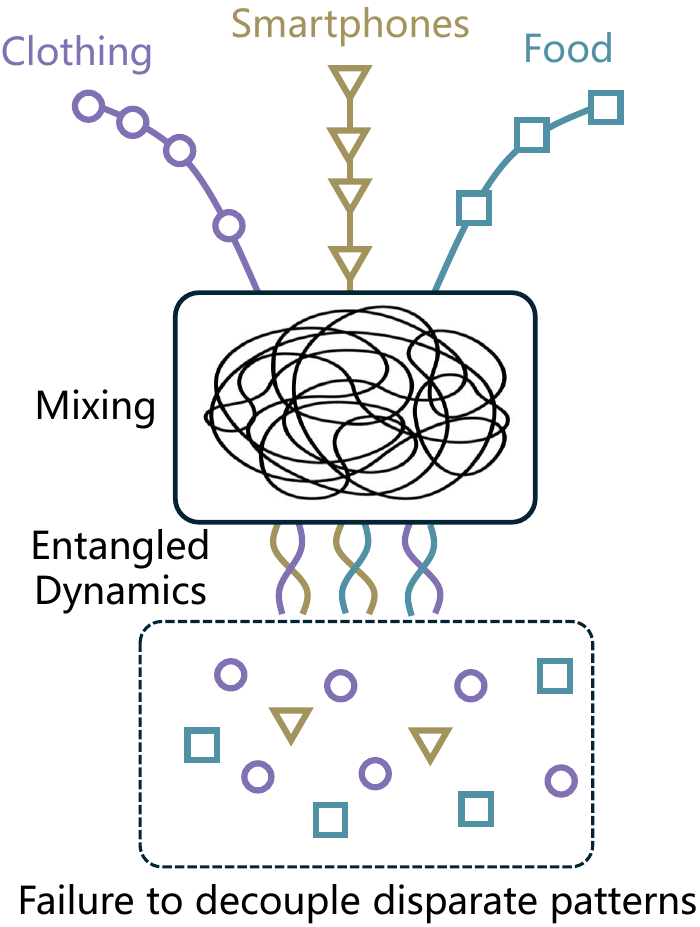}\label{fig:channel-mixing}}
\hfil
\subfloat[Channel independence]{\includegraphics[height=2.1in]{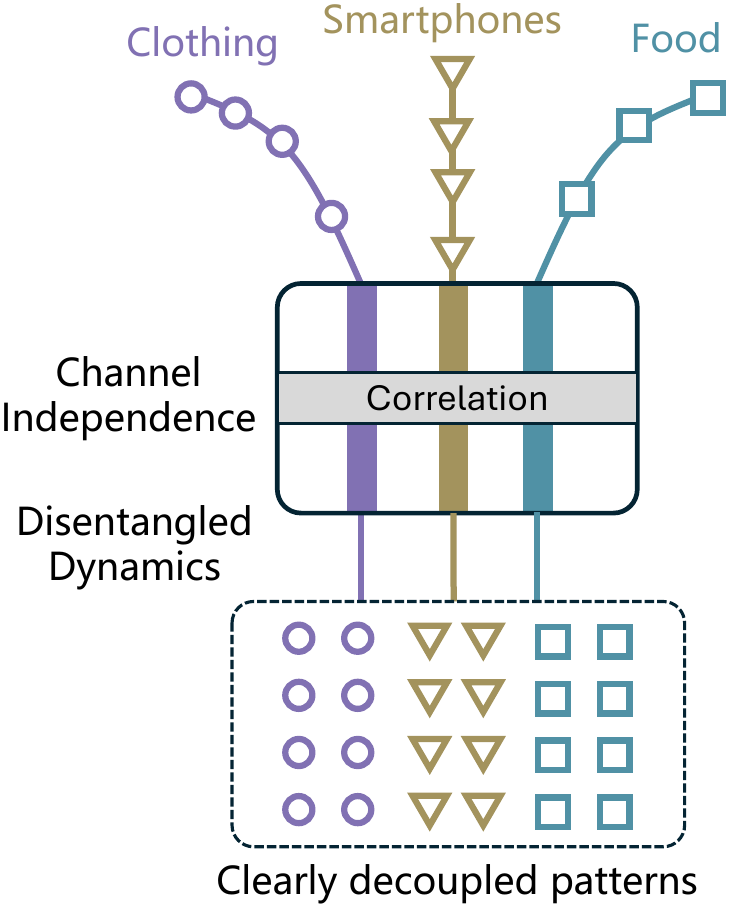}\label{fig:channel-independence}}
\caption{An illustration of channel-mixing versus channel-independent architectures applied to user browsing history from the Taobao e-commerce dataset. The channel-independent approach avoids the pattern confusion inherent in channel-mixing.}
\label{fig:intro}

\end{figure}

\begin{figure*}
    \centering
    \includegraphics[width=0.8\linewidth]{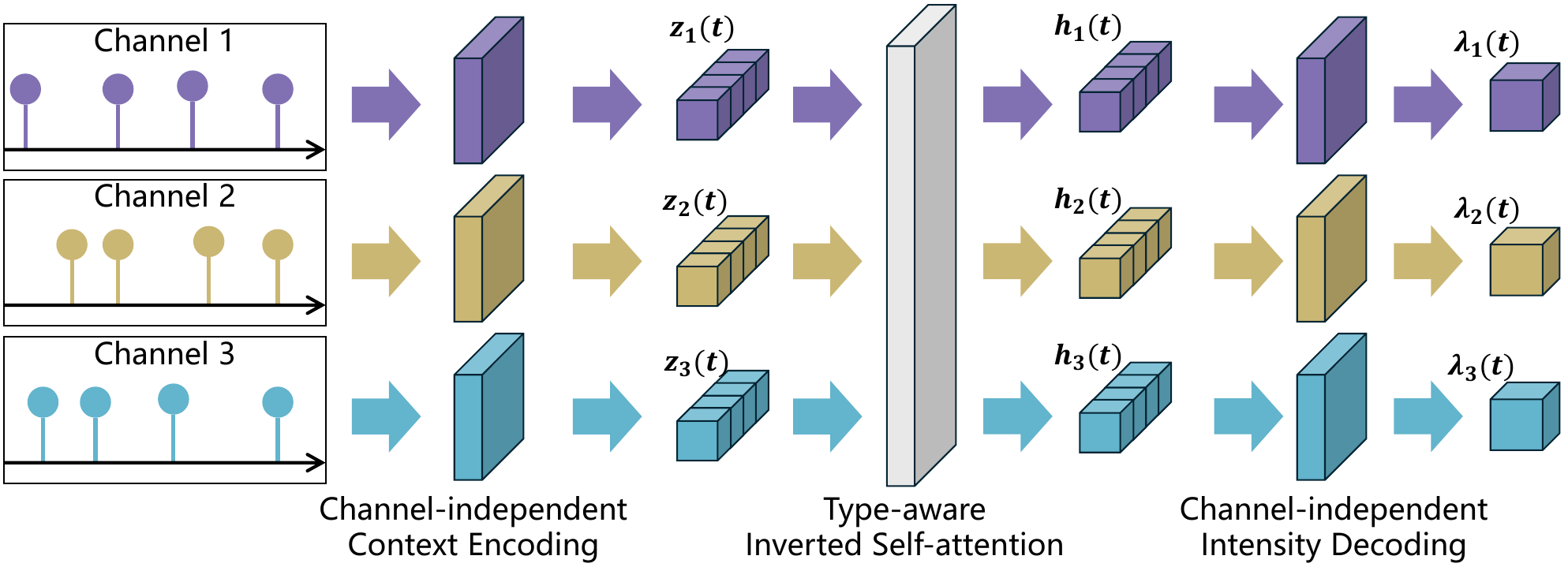}
    \caption{The general architecture of the proposed ITPP model. The framework has a channel-independent encoder-decoder architecture, with a type-aware inverted self-attention layer lying in the middle, which explicitly captures the latent correlations between different channels. }
    \label{fig:architecture}
\end{figure*}

Existing MTPP models predominantly adopt a channel-mixing approach, encoding the entire history of diverse event types into a unified context embedding. Fig. \ref{fig:channel-mixing} illustrates the problem of channel-mixing approaches. By forcing a single representation to encapsulate the complex and often disparate dynamics of all event types, e.g., browsing history of irrelevant products, the model's ability to capture the unique, type-specific temporal patterns crucial for accurate prediction can be diluted. This entanglement of information may introduce noise and interference. Consequently, the model may struggle to distinguish and learn the distinct generative processes of each event type, leading to a suboptimal understanding of the underlying dynamics and ultimately, a compromised predictive performance. For instance, when the dynamics of user browsing history for disparate categories like clothing, smartphones, and food are combined into an entangled representation, the model may struggle to isolate category-specific patterns. As a result, predictions for one category can be polluted by noise from another, causing a decline in overall accuracy.


To address this, we propose a strategy that first decoupling event dynamics and then explicitly models inter-channel dependencies. This channel-independent strategy is illustrated in Fig. \ref{fig:channel-independence}. We begin by decomposing event sequences into distinct channels, each containing the dynamics regarding a specific event type. This parallel simulation of each channel's state evolution prevents interference between heterogeneous patterns. Recognizing that inter-type correlations are critical for accurate prediction, we then introduce a correlation layer, which is designed specifically to capture the reciprocal effects among channels, e.g., one event type may excite or inhibit another. This channel-independent strategy guarantees a set of decoupled representations for disparate dynamic patterns of different event types. We argue that the proposed architecture enables the model to learn a complete and robust representation of the system's dynamics.

This paper introduces the channel-\textbf{I}ndependent marked \textbf{T}emporal \textbf{P}oint \textbf{P}rocess (ITPP), a model designed to enhance fine-grained event prediction by disentangling type-specific event dynamics while explicitly modeling inter-type correlations. As illustrated in Fig. \ref{fig:architecture}, ITPP employs a novel `encoding-correlation-decoding' architecture. The encoding and decoding stages operate independently on each channel, while an interposed inverted self-attention layer captures cross-channel dependencies. Extensive experiments demonstrate the superiority of our model over state-of-the-art methods. The contributions of this paper can be summarized as follows:
\begin{itemize}
\item We propose a novel channel-indepedent architecture for MTPP modelling, named ITPP, which adopts an encoder-decoder architecture with an ODE-based backbone. This framework disentangles the heterogeneous dynamics of different event types, which we argue helps improve the robustness of the model.

\item We devise a type-aware inverted self-attention module that captures the correlations between different event channels. A set of type-aware biases is integrated with the vanilla attention mechanism to preserve the inherent connections between different event types while incorporating state-based correlations.

\item Extensive experiments are conducted to validate the superiority of the proposed ITPP framework over state-of-the-art MTPP models. 
\end{itemize}

\section{Related works}
\subsection{Marked temporal point processes}
Most Marked Temporal Point Process (MTPP) models use an encoder-decoder architecture to handle asynchronous event data. In this framework, an encoder summarizes the event history into a fixed-size embedding, which a decoder then uses to predict the distribution of the next event. The majority of prior research has concentrated on designing more powerful encoders. 
\cite{rmtpp,nhp} extends RNNs to a sophisticated continuous-time version to jointly encode event marks and irregular time intervals.
 \cite{SAHP,THP,attnhp,eventformer} propose modified versions of self-attention modules integrated with elaborately designed temporal embeddings, to replace RNNs as the sequential encoding backbone. Other architectures like CNNs \cite{ctpp} and MLPs \cite{FullyNN} have also been proved effective in MTPP encoding.
Recently, neural ordinary differential equations \cite{neuralode,gruode} have become an effective tool for modelling infinitesimal dynamic evolution. \cite{njsde,nstpp,gstpp,njdtpp} adopt ODE-based architectures for encoders, where latent states are formulated as fine-grained trajectories that evolve over time. Some other works devise different decoding architectures other than intensity fitting. \cite{lognormmix} propose to use a mixture of log normal distributions to fit the target distribution. 
Denoising diffusion models have also been applied to MTPP in more complex scenarios such as long-term event prediction \cite{dltpp}, spatio-temporal event modelling \cite{dstpp} and high-dimensional event prediction \cite{high-dim}. Existing MTPP models generally use a channel-mixing approach, in which information from different event types is mixed into a single encoding vector. We argue that this entanglement of information undermines the model's robustness and potentially lead to overfitting.

\subsection{Channel independence}
The concept of channel independence has been explored in multivariate time series forecasting.  \cite{patchtst} propose to model each variable of a multivariate time series independently with Transformers. This approach stands in contrast to traditional channel-mixing models, which immediately project all variables into a shared latent space.
 They argue that the most valuable predictive information is contained within the history of a channel itself, and that premature mixing can corrupt these signals. 
This idea is pushed further by \cite{tsmixer,dlinear,filternet,frequency}, which demonstrate the power of this principle under different architectures. Recently, iTransformer \cite{itransformer} has attracted a great deal of attention by combining a channel-independent encoding architecture and an attentive correlation learning module. It treats the entire time series of individual variates as tokens and applies attention across the different independent channels. Despite the demonstrated success of channel independence in time series forecasting, to our knowledge, this principle has not yet been extended to MTPP modeling, a domain where information disentanglement is arguably even more critical. MTPP modelling differs greatly from multivaraite time series forecasting in terms of asynchronization and semantic discretization, and therefore existing frameworks for channel-independence are not applicable in this scenario. In this work, we reformulate the idea of channel-independence for MTPP modelling to decouple disaprate dynamics of different event types.

\section{Preliminary: marked temporal point processes}
A Marked Temporal Point Process (MTPP) models a sequence of events occurring in continuous time, where each event $i$ is a time-mark pair $(t_i, m_i)$. The events are ordered by their arrival times $0 < t_1 \le t_2 \le \dots \le t_L < T$, and each mark $m_i$ belongs to a discrete set of $K$ event types.
The behavior of an MTPP is fully characterized by its conditional intensity function, $\lambda_m(t|\mathcal{H}_t)$, where $\mathcal{H}_t = \{(t_j, m_j) | t_j < t\}$ is the history of past events. For clarity, the history condition $\mathcal{H}_t$ is omitted from our notation for the remainder of this paper.
 The intensity function represents the instantaneous rate of an event occurrence, defined as the expectation of the number of event occurrences per unit of time:
\begin{equation}
    \lambda(t)=\lim_{\Delta t\rightarrow 0}\frac{\mathbb{E}[N(t+\Delta t)-N(t)]}{\Delta t}
\end{equation}
where $ N(t) $ is a counting measure. A key advantage of the conditional intensity function is that it only needs to be non-negative, unlike a probability density function (PDF) which must integrate to one. This modeling flexibility has led most MTPP research to focus on directly parameterizing the intensity function. 


\section{Methodology}
In this section, we introduce the proposed ITPP framework, which employs channel-independent encoding and decoding to preserve type-specific dynamic patterns, while adopting a type-aware inverted self-attention mechanism to capture the correlation between different event channels. The general architecture of the proposed model is shown in Fig. \ref{fig:architecture}.

\subsection{Model architecture}
As shown in Fig. \ref{fig:architecture}, the ITPP model can be divided into three components, namely channel-independent context encoding, type-aware inverted self-attention and channel-independent intensity decoding. Inspired by recent works like \cite{patchtst}, we propose to treat event occurrences of different types as separate time series and group them into separate channels. Each channel is then encoded independently into a time-variable context embedding $ \boldsymbol{z}_k(t)\in \mathbb{R}^{d} $, where $ k\in\{1, ...,K\} $ is the channel index and $ t $ is the underlying timestamp. Then, an inverted self-attention layer is applied to exploit the explicit correlation between different channels. Unlike the vanilla Transformers \cite{transformer} or iTransformers \cite{itransformer}, where mapping parameters are generally shared among different entries, our type-aware inverted self-attention uses channel-specific parameters in order to preserve natural connections between event types rather than only state-dependent correlations. The attention output $ \boldsymbol{h}_k(t)\in \mathbb{R}^d $ is finally independently decoded to get the type-specific intensity value $ \lambda_k(t) $. 

\subsection{Channel-independent context encoding}
\begin{figure}[tb]
    \centering
    \includegraphics[width=.9\linewidth]{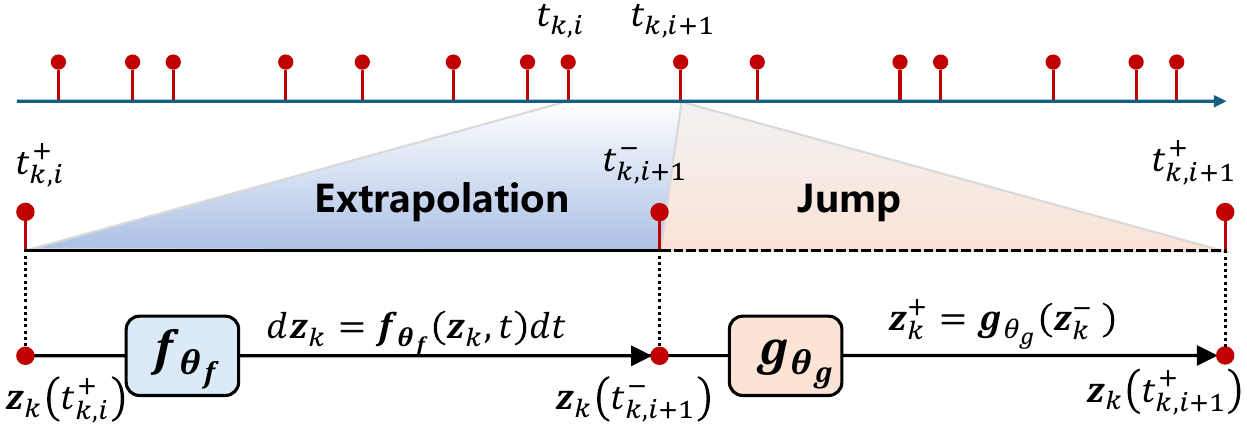}
    \caption{The architecture of channel-independent context encoding. The encoder simulates two types of state transition, namely extrapolation and jump.}
    \label{fig:encoding}
\end{figure}

Unlike state-of-the-art MTPP models, which adopt channel-mixing approaches to encode the history event sequence into a single hidden state, we separate the sequence into multiple channels, each representing an event type. We adopt neural ODEs with jumps to model the fine-grained dynamics of the temporal context. Different channels are simulated simultaneously with the same model formulation and shared parameters. The architecture of our proposed channel-independent context encoding is shown in Fig. \ref{fig:encoding}.
The channel-independent state evolution can be divided into two types, namely extrapolations and jumps. The extrapolation process is used to model the smooth state transition within event intervals, while a jump simulates the abrupt state change induced by an event occurrence. Specifically, given the posterior hidden state of the $ i $-th event in the $ k $-th channel, denoted as $ \boldsymbol{z}_k(t_{k,i}^+) $, the extrapolation process can be defined as the following ODE:
\begin{equation}
    d\boldsymbol{z}_k(t)=\boldsymbol{f}_{\boldsymbol{\theta}_f}(\boldsymbol{z}_k(t),t)dt
\end{equation}
where $ \boldsymbol{f}_{\boldsymbol{\theta}_f} $ is the drift function, parameterised by $ {\boldsymbol{\theta}_f} $. The hidden state of any time before the next event occurrence can be computed by solving an initial value problem:
\begin{equation}
    \boldsymbol{z}_k(t)=\boldsymbol{z}_k(t_{k,i}^+)+\int_{t_{k,i}^+}^t\boldsymbol{f}_{\boldsymbol{\theta}_f}(\boldsymbol{z}_k(\tau),\tau)d\tau
\end{equation}
where $ t_{k,i}^+\leq t\leq t_{k,i+1}^+ $. A jump is parametrised with a state update neural network. Given the prior state of the $ i $-th event in the $ k $-th channel, denoted as $ \boldsymbol{z}_k(t_{k,i}^-) $, the jump is defined as:
\begin{equation}
    \boldsymbol{z}_k(t_{k,i}^+)=\boldsymbol{g}_{\boldsymbol{\theta}_g}(\boldsymbol{z}_k(t_{k,i}^-))
\end{equation}
where $ \boldsymbol{g}_{\boldsymbol{\theta}_g} $ is the jump function, parameterised by $ {\boldsymbol{\theta}_g} $. The encoded context hidden state $ \boldsymbol{z}_k(t) $ is then fed to the type-aware inverted self-attention layer to capture inter-channel correlations, which will be introduced in the next subsection.

\subsection{Type-aware inverted self-attention}
\begin{figure}[tb]
    \centering
    \includegraphics[width=0.5\linewidth]{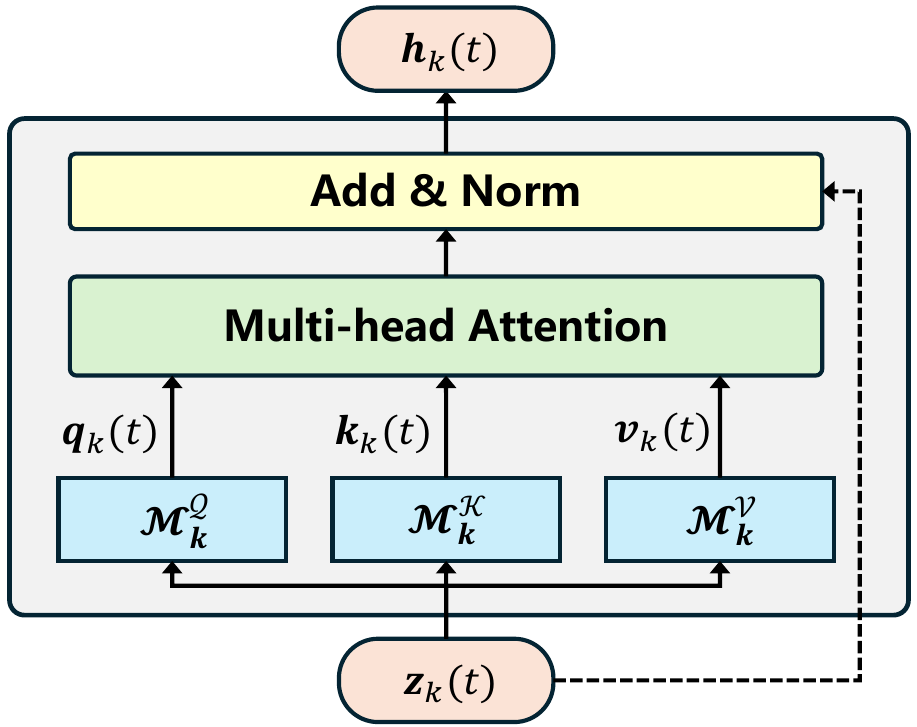}
    \caption{The architecture of type-aware inverted self-attention. The property of type-awareness is achieved by channel-specific mapping functions.}
    \label{fig:iAttention}
\end{figure}

The type-aware inverted self-attention layer, parameterised by $ {\boldsymbol{\theta}_c} $, is used to capture the latent correlations between different event channels. The general architecture of this module is shown in Fig. \ref{fig:iAttention}. 
\begin{table*}[!ht]
  \setlength{\tabcolsep}{0pt}
  \begin{tabular*}{\textwidth}{@{\extracolsep{\fill}}l *{12}{c}}
    \toprule
    & \multicolumn{3}{c}{\texttt{StackOverflow}} & \multicolumn{3}{c}{\texttt{MIMIC}} & \multicolumn{3}{c}{\texttt{Taobao}} & \multicolumn{3}{c}{\texttt{Earthquake}} \\
    \cmidrule(lr){2-4} \cmidrule(lr){5-7} \cmidrule(lr){8-10} \cmidrule(lr){11-13}
    Methods & TM-NLL & T-NLL & M-NLL & TM-NLL & T-NLL & M-NLL & TM-NLL & T-NLL & M-NLL & TM-NLL & T-NLL & M-NLL \\
    \midrule
    RMTPP       & 2.303 & 0.724 & 1.580 & 1.801 & 0.673 & 1.128 & -0.113 & -1.510 & 1.407 & 1.819 & 0.474 & 1.346 \\
    NHP         & 2.233 & 0.703 & \textit{1.531} & 1.204 & \textit{0.451} & 0.753 & -1.003 & -2.500 & 1.497 & 1.724 & 0.382 & \underline{1.342} \\
    LogNormMix  & \underline{2.198} & \textbf{0.626} & 1.572 & \textit{1.191} & \underline{0.430} & 0.761 & \underline{-1.176} & \underline{-2.539} & \underline{1.363} & \underline{1.706} & \underline{0.367} & \textbf{1.339} \\
    THP         & 2.325 & 0.772 & 1.553 & 1.600 & 0.601 & 0.999 & 0.260  & -1.337 & 1.597 & 1.906 & 0.561 & 1.345 \\
    SAHP        & 2.268 & 0.705 & 1.562 & 1.278 & 0.495 & 0.783 & -0.999 & -2.495 & 1.497 & 1.740 & 0.389 & 1.351 \\
    NeuralODE   & 2.251 & 0.657 & 1.594 & 1.352 & 0.477 & 0.874 & -1.039 & -2.496 & 1.457 & 1.722 & \textit{0.371} & 1.351 \\
    ODE-GRU     & \textit{2.228} & 0.715 & \underline{1.514} & 1.340 & 0.463 & 0.877 & -0.888 & -2.376 & 1.488 & 1.731 & 0.388 & 1.343 \\
    CTPP        & 2.229 & \textit{0.651} & 1.578 & \underline{1.176} & 0.464 & \underline{0.713} & \textit{-1.158} & \textbf{-2.570} & \textit{1.412} & \textit{1.715} & 0.373 & \underline{1.342} \\
    AttNHP      & 2.267 & 0.735 & 1.532 & 1.438 & 0.595 & 0.842 & -0.969 & -2.401 & 1.432 & 1.717 & 0.374 & 1.344 \\
    NJDTPP      & 2.398 & 0.772 & 1.626 & 1.374 & 0.643 & \textit{0.731} & -0.439 & -2.004 & 1.565 & 1.802 & 0.451 & 1.351 \\
    \midrule
    \textbf{ITPP} & \textbf{2.106} & \underline{0.636} & \textbf{1.470} & \textbf{1.054} & \textbf{0.427} & \textbf{0.627} & \textbf{-1.189} & \textit{-2.538} & \textbf{1.349} & \textbf{1.692} & \textbf{0.347} & 1.344 \\
    \bottomrule
  \end{tabular*}
  \caption{Probabilistic evaluation results, with the first, second and third best ones shown in bold, underlined and italic styles, respectively.}
  \label{tab:nll}
\end{table*}
This layer adopts a similar structure as the vanilla Transformer. However, unlike other application areas such as Natural Language Processing (NLP), Computer Vision (CV), and time series forecasting, where entries (word tokens, pixels, time steps, etc.) are homogeneous items with positional relations, the channels in our scenario have distinct nature from each other--different event types should not be considered equal entries in the attention process. Thus, we use different parameterisations of the linear mapping layer for different channels. Specifically, the query, key and value computation of channel $ k $ is defined as follows:
\begin{align}
    \boldsymbol{q}_k(t)=\boldsymbol{\mathcal{M}}_k^{\mathcal{Q}}(\boldsymbol{z}_k(t))&=\boldsymbol{W}^{\mathcal{Q}}\boldsymbol{z}_k(t)+\boldsymbol{b}^{\mathcal{Q}}_k\\
    \boldsymbol{k}_k(t)=\boldsymbol{\mathcal{M}}_k^{\mathcal{K}}(\boldsymbol{z}_k(t))&=\boldsymbol{W}^{\mathcal{K}}\boldsymbol{z}_k(t)+\boldsymbol{b}^{\mathcal{K}}_k\\
    \boldsymbol{v}_k(t)=\boldsymbol{\mathcal{M}}_k^{\mathcal{V}}(\boldsymbol{z}_k(t))&=\boldsymbol{W}^{\mathcal{V}}\boldsymbol{z}_k(t)+\boldsymbol{b}^{\mathcal{V}}_k
\end{align}
where $ \boldsymbol{W}^{\mathcal{Q}} $, $ \boldsymbol{W}^{\mathcal{K}} $ and $ \boldsymbol{W}^{\mathcal{V}} $ are the mapping matrices shared across different channels, while $ \boldsymbol{b}^{\mathcal{Q}}_k $, $ \boldsymbol{b}^{\mathcal{K}}_k $ and $ \boldsymbol{b}^{\mathcal{V}}_k $ are channel-specific biases that preserves inherent latent features of event types. The cross-channel self-attention is then performed as:
\begin{equation}
    \mathrm{ATTN}(\boldsymbol{Q}(t),\boldsymbol{K}(t),\boldsymbol{V}(t))=\mathrm{softmax}(\frac{\boldsymbol{Q}(t)\boldsymbol{K}^T(t)}{\sqrt{d_k}})\boldsymbol{V}(t)
\end{equation}
where $ \boldsymbol{Q}(t)\in \mathbb{R}^{K\times d_k} $, $ \boldsymbol{K}(t)\in \mathbb{R}^{K\times d_k} $ and $ \boldsymbol{V}(t)\in \mathbb{R}^{K\times d_v} $ are the query, key and value matrices whose rows are stacked by $ \boldsymbol{q}_k(t) $, $ \boldsymbol{k}_k(t) $ and $ \boldsymbol{v}_k(t) $ respectively. Following the design of the vanilla Transformer, residual connection and layer normalisation are also adopted for stable performance.

\subsection{Channel-independent intensity decoding}
The intensity decoding layer computes the intensity value $ \lambda_k(t) $ from the hidden state $ \boldsymbol{h}_k(t) $. The decoding process is defined as follows:
\begin{equation}
    \lambda_k(t)=\boldsymbol{r}_{\boldsymbol{\theta}_r}(\boldsymbol{h}_k(t))
\end{equation}
where $ \boldsymbol{r}_{\boldsymbol{\theta}_r} $ is the channel-independent decoding network with shared parameters $ {\boldsymbol{\theta}_r} $ across channels. The joint probability of the next event can be calculated as:
\begin{equation}
    f(t,k)=\lambda_k(t)\exp(-\int_{\Bar{t}}^t\lambda(\tau)d\tau)
\end{equation}
where $ \Bar{t} $ is the arrival time of the last known event and $ \lambda(t)=\sum_k \lambda_k(t) $ is the total intensity of all event types.

\subsection{Training}
We train our model using the commonly adopted Maximum Log-likelihood Estimation (MLE) approach. The training loss is the Negative Log-Likelihood (NLL) of the event sequence $ \mathcal{S}=\{(t_i,m_i)\}_{i=1}^L $ in a rolling prediction manner as follows:
\begin{equation}
    \mathcal{L}_{\boldsymbol{\theta}}(\mathcal{S})=-\sum_{i=1}^L\log \lambda_{m_i}(t_i)+\int_0^T \lambda(\tau)d\tau
\end{equation}
where $ \boldsymbol{\theta}=\{\boldsymbol{\theta}_f,\boldsymbol{\theta}_g,\boldsymbol{\theta}_c,\boldsymbol{\theta}_r\} $ are the trainable model parameters. The gradients of the model parameters are computed by backpropagation with the help of the adjoint sensitivity method \cite{neuralode}, which runs the ODEs backward in time.


\begin{table}[t]
  \setlength{\tabcolsep}{2pt}
  \begin{tabular*}{\columnwidth}{@{\extracolsep{\fill}}lcccc}
    \toprule
    Datasets      & \textbf{\# events} & \textbf{Avg len.} & \textbf{\# seqs} & \textbf{\# types} \\
    \midrule
    StackOverflow & 142,777            & 65                    & 2,203                 & 22                \\
    MIMIC         & 2,419              & 4                     & 650                   & 75                \\
    Taobao        & 115,397            & 58                    & 2,000                 & 17                \\
    Earthquake    & 70,723             & 16                    & 4,300                 & 7                 \\
    \midrule
    Poisson       & 39,893             & 80                    & 500                   & 3                 \\
    Hawkes        & 21,166             & 42                    & 500                   & 3                 \\
    \bottomrule
  \end{tabular*}
  \caption{Dataset statistics}
    \label{tab:dataset}
\end{table}

\section{Experiments}
To comprehensively evaluate our proposed model, ITPP, we conduct four sets of experiments, namely probabilistic evaluation, prediction evaluation, intensity recovery and ablation study. Collectively, these experiments validate the effectiveness and robustness of ITPP. Code is available online\footnote{https://github.com/AnthonyChouGit/ITPP}.

\subsection{Datasets}
We evaluate our model on six datasets, comprising four real-world benchmarks—StackOverflow \cite{stackoverflow}, MIMIC \cite{mimic}, Taobao \cite{hypro}, and Earthquake \cite{easytpp}—and two synthetic datasets generated from a Poisson and a Hawkes process, respectively. These datasets vary in scale, sequence length, and number of event types. The statistics of the datasets are given in Table \ref{tab:dataset}. 
This heterogeneity allows for a rigorous assessment of our model's robustness and its ability to accommodate across varied data characteristics, which we will compare against state-of-the-art methods. Please refer to the Appendix for more information regarding these datasets.

\begin{figure*}[t]
    \centering
    \includegraphics[width=.8\linewidth]{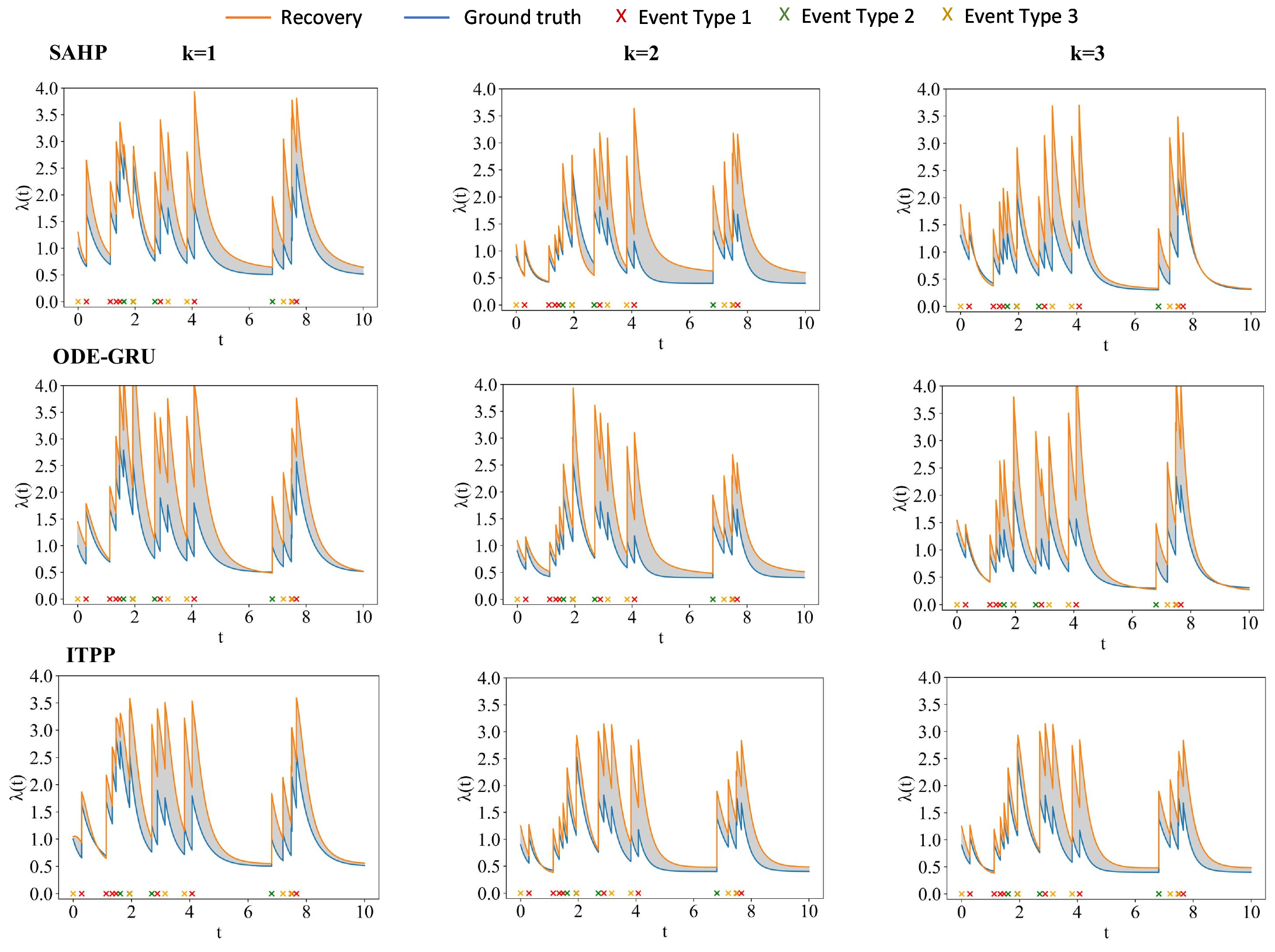}
    \caption{Visualisation of intensity recovery. The grey areas indicate the mass of recovery loss.}
    \label{fig:vis}
\end{figure*}

\subsection{Baselines}
For a comprehensive comparison, we evaluate our model against 10 baselines. These methods span several architectural categories, including RNN-based (RMTPP \cite{rmtpp}, NHP \cite{nhp}, LogNormMix \cite{lognormmix}), self-attentive (THP \cite{THP}, SAHP \cite{SAHP}, AttNHP \cite{attnhp}), convolutional (CTPP \cite{ctpp}), and differential equation-based models (NeuralODE \cite{neuralode}, ODE-GRU \cite{gruode}, NJDTPP \cite{njdtpp}). A brief introduction of each baseline is given in the Appendix.


\begin{table}[t]
  \setlength{\tabcolsep}{0pt}
  \begin{tabular*}{\columnwidth}{@{\extracolsep{\fill}}lcccccc}
    \toprule
    & \multicolumn{2}{c}{\texttt{SO}} & \multicolumn{2}{c}{\texttt{MIMIC}} & \multicolumn{2}{c}{\texttt{Earth.}} \\
    \cmidrule(lr){2-3} \cmidrule(lr){4-5} \cmidrule(lr){6-7}
    Methods & RMSE & F1 & RMSE & F1 & RMSE & F1 \\
    \midrule
    RMTPP     & 1.021 & 0.306 & 0.888 & 0.823 & 1.249 & 0.317 \\
    NHP       & 1.017 & 0.299 & 0.863 & \underline{0.850} & \underline{1.231} & 0.303 \\
    SAHP      & 1.019 & 0.299 & \underline{0.842} & 0.834 & 1.252 & 0.309 \\
    NeuralODE & 1.020 & 0.293 & 0.884 & 0.847 & 1.238 & 0.328 \\
    ODE-GRU   & \underline{1.009} & \textbf{0.313} & 0.873 & 0.836 & 1.252 & 0.307 \\
    NJDTPP    & 1.030 & 0.321 & 0.957 & 0.841 & 1.255 & \textbf{0.320} \\
    \midrule
    \textbf{ITPP} & \textbf{1.007} & \textbf{0.313} & \textbf{0.831} & \textbf{0.856} & \textbf{1.226} & \underline{0.319} \\
    \bottomrule
  \end{tabular*}
  \caption{Prediction evaluation results}
    \label{tab:pred}
\end{table}

\subsection{Probabilistic evaluation}
\label{sec:nll}
We evaluate ITPP's probabilistic performance against state-of-the-art models on four real-world datasets, using the average negative log-likelihood (NLL) per event. The primary metric, the joint NLL of arrival times and marks (TM-NLL), quantifies the models' overall fitting capability. As shown in Table \ref{tab:nll}, ITPP achieves the best TM-NLL results on all datasets, surpassing the second-best models by a significant margin in every case.

A comparison across different model architectures yields some interesting insights. First,  contrary to expectations, self-attentive models (THP, SAHP, and AttNHP) do not outperform strong RNN-based counterparts like LogNormMix and NHP, even on datasets with long sequences (StackOverflow and Taobao) where their ability to capture long-range dependencies should be most advantageous. We hypothesize that this is because self-attention mechanisms inherently lack a strong representation of sequential and temporal information, a factor that is more fundamental to event prediction tasks than to typical NLP applications. Second, the performance of ODE-based models like NeuralODE and ODE-GRU is inconsistent, showing a competitive probabilistic fit only on the StackOverflow dataset, which contains the largest number of events (see Table \ref{tab:dataset}). This suggests a vulnerability to overfitting, as their continuous dynamics may struggle to disentangle information from multiple event types without a large amount of data. However, ITPP resolves this problem with a channel-independent strategy, and achieves a significant performance boost, proving the effectiveness of ODE-based architecture in MTPP modelling.

The time (T-NLL) and mark (M-NLL) components reveal a notable trade-off in the models' focus. On datasets with a relatively large number of event types (StackOverflow, MIMIC, and Taobao, see Table \ref{tab:dataset}), ITPP exhibits significant performance improvements in M-NLL, at the expense of a slight decrease in T-NLL performance. Conversely, on the Earthquake dataset, which has few event types, it focuses more on time loss. We attribute this behavior to ITPP's channel-independent design, which disentangles the dynamics of different event types. This separation enables the model to better distinguish between marks, making it much easier to reduce the mark loss. This effect becomes more pronounced as the number of event types increases, thereby shifting the model's focus to the mark component.

\begin{figure}[h]

\centering
\subfloat[Poisson]{\includegraphics[width=0.45\linewidth]{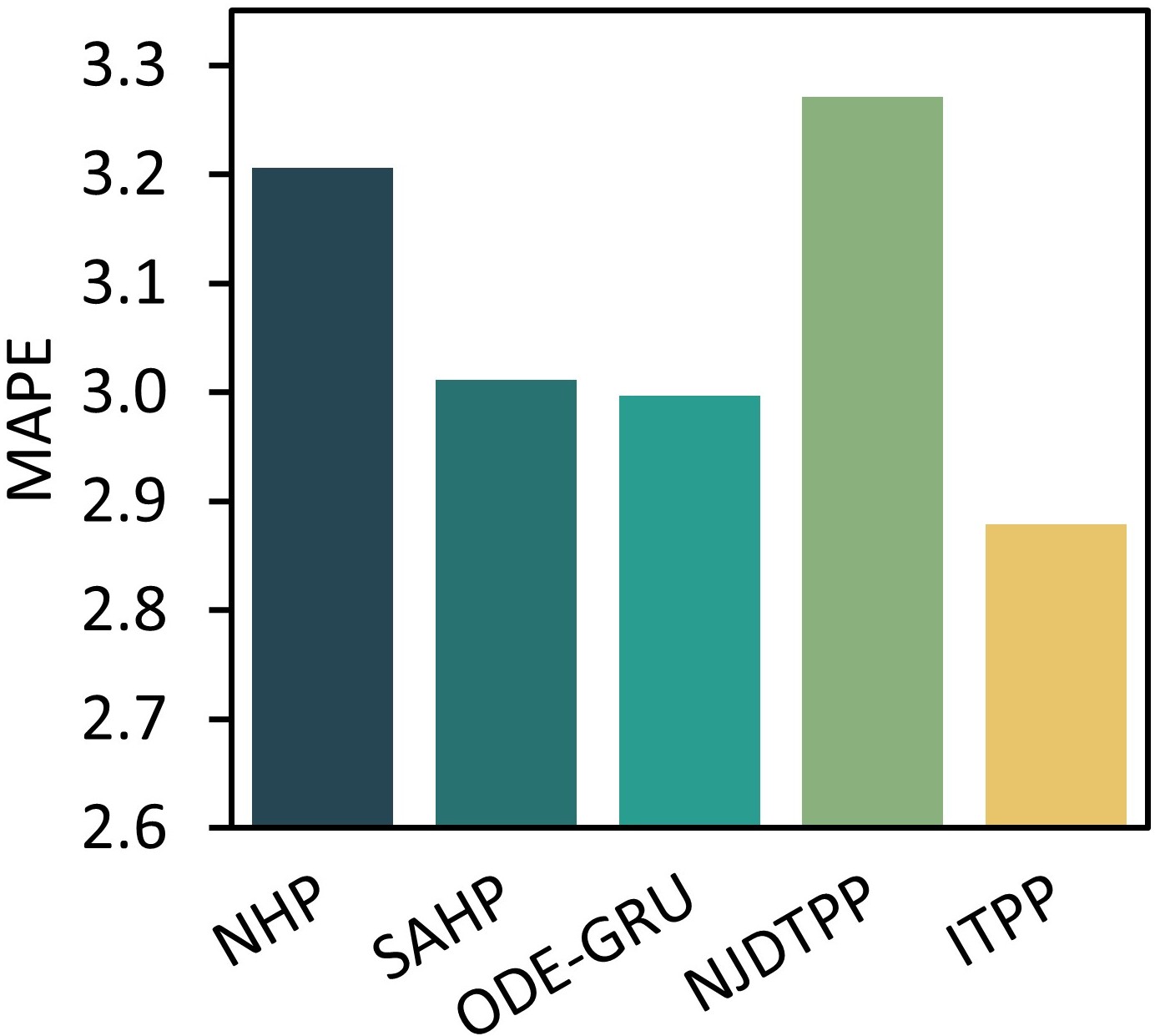}}
\hfil
\subfloat[Hawkes]{\includegraphics[width=0.45\linewidth]{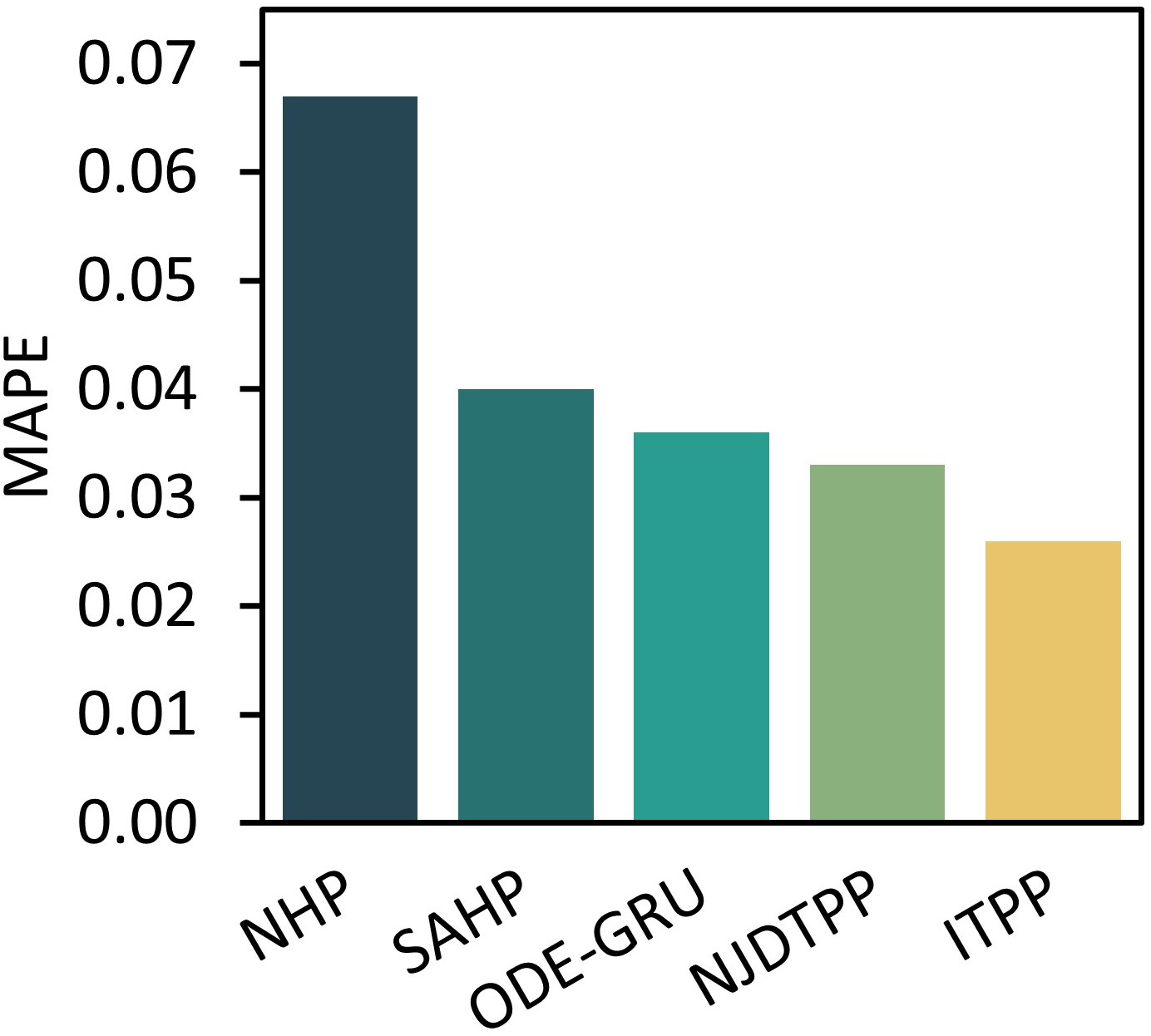}}
\caption{Mean absolute percentage error of intensity recovery on synthetic datasets.}
\label{fig:mape}

\end{figure}

\subsection{Prediction evaluation}
\label{sec:pred}
We now assess the models' capabilities on prediction tasks using three real-world datasets. 
The Root Mean Square Error (RMSE) and F1 score are used as the metrics for the evaluation of time and mark prediction, respectively. The prediction evaluation results are shown in Table \ref{tab:pred}. 
Overall, ITPP dominates the prediction tasks, securing top performance on five of the six indicators and falling only marginally short on the F1 score for the Earthquake dataset. This robust performance contrasts with the inconsistency of the channel-mixing ODE-GRU, which—mirroring the probabilistic results—performs well only on the extensive StackOverflow dataset. This pattern reinforces the conclusion that channel-mixing ODE architectures struggle with generalization and are susceptible to overfitting without large-scale data.

\subsection{Intensity recovery}
This section evaluates the models' ability to recover the underlying conditional intensity function. We use two synthetic datasets, Poisson and Hawkes, for this analysis, as their ground-truth intensity functions are analytically known. The Mean Absolute Percentage Error (MAPE) is used to measure performance, with the results presented in Figure \ref{fig:mape}.
ITPP significantly outperforms existing intensity-based MTPP models in intensity recovery on both synthetic datasets. This provides a clear rationale for ITPP's superior performance in the probabilistic fitting and point prediction tasks discussed previously. 
For a qualitative view, Figure \ref{fig:vis} visualizes the recovered intensities from the top three models on the Hawkes dataset. The plot confirms the quantitative results, illustrating that ITPP's predicted intensity more closely tracks the ground-truth for all three event types, achieving a visibly tighter fit.

\subsection{Ablation study}
We now conduct an ablation study to quantify the specific contributions of our two core design choices: channel independence and inverted self-attention. The following experiments are designed to demonstrate that both concepts are crucial to the model's superior performance.

\subsubsection{Channel independence}
\begin{figure}[tb]

\centering
\subfloat[Stackoverflow]{\includegraphics[width=0.45\linewidth]{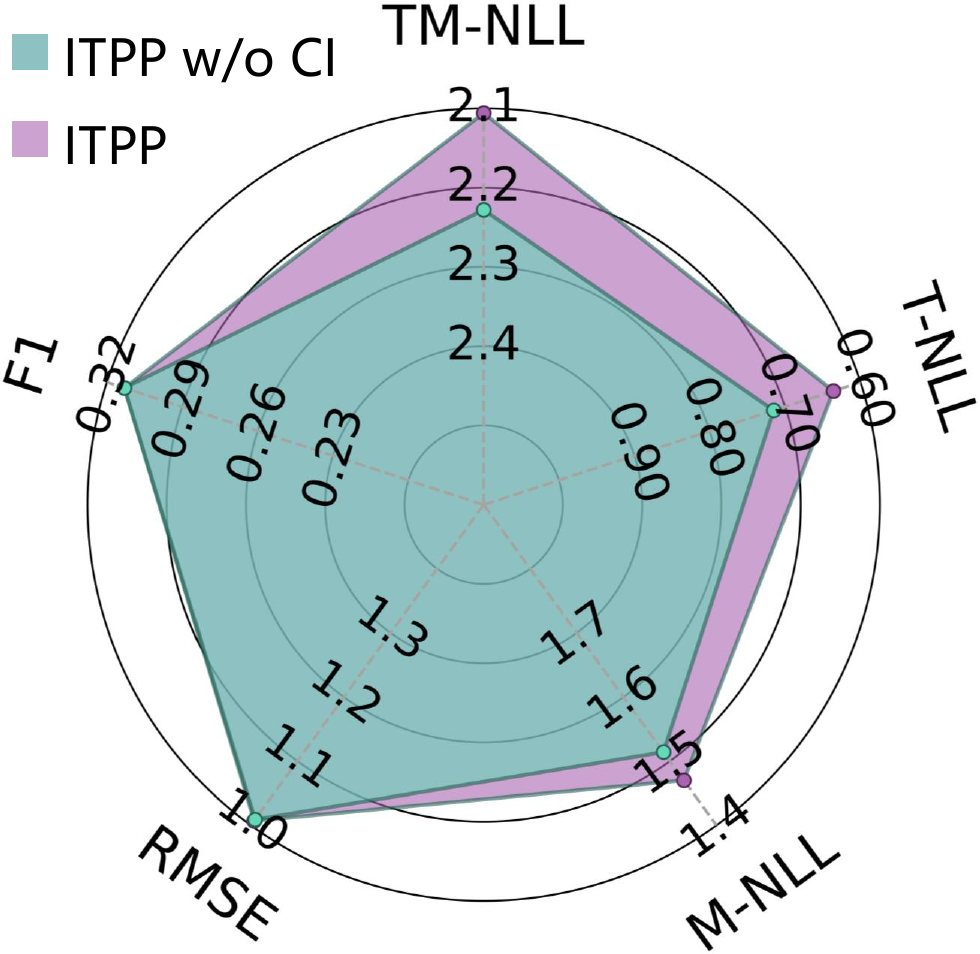}}
\hfil
\subfloat[Earthquake]{\includegraphics[width=0.45\linewidth]{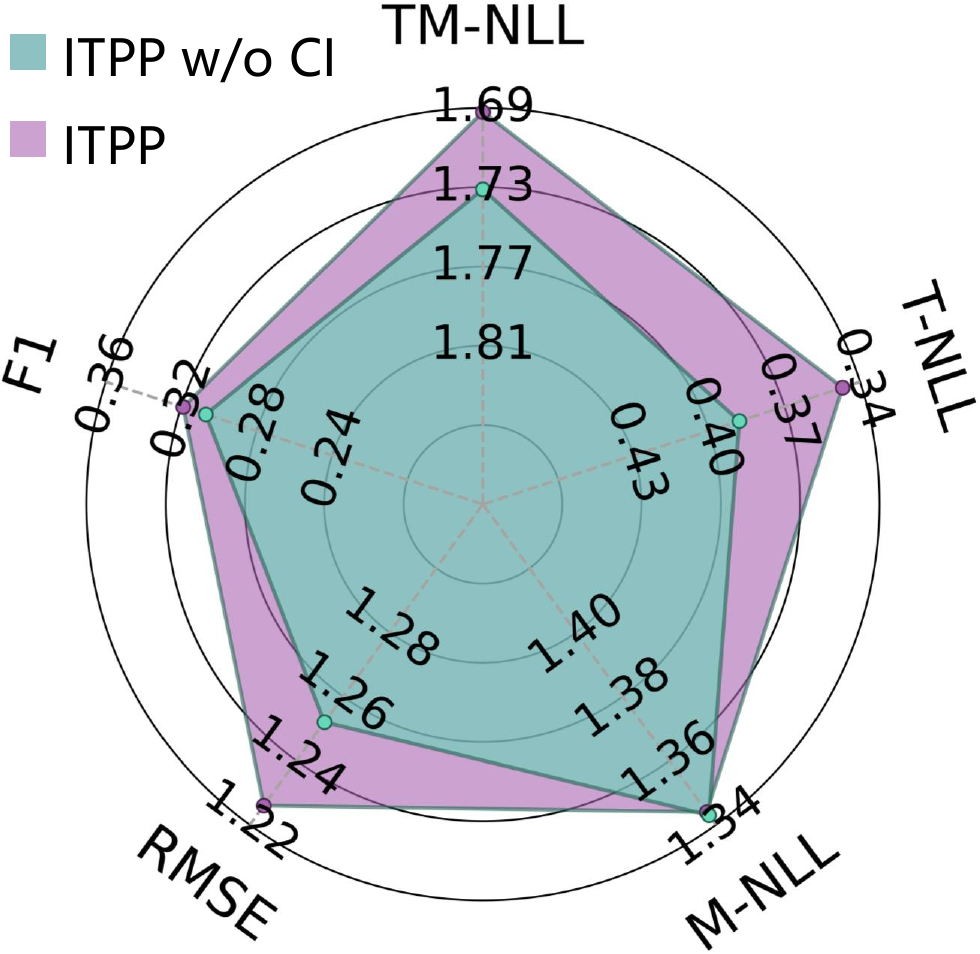}}
\caption{Ablation study results of channel independence.}
\label{fig:ablation-ci}

\end{figure}

To quantify the impact of our channel-independent design, we compare the full ITPP model against ITPP w/o CI, a variant that reverts to a conventional channel-mixing approach. The evaluation is conducted across the five metrics used in our previous probabilistic and prediction analyses. Fig. \ref{fig:ablation-ci} shows the comparison results on StackOverflow and Earthquake datasets.
The results demonstrate that the channel-independent design provides a significant performance improvement. This effect is particularly pronounced on smaller-scale datasets, such as Earthquake, as shown in the figure.

\subsubsection{Inverted self-attention}
\begin{figure}[t]

\centering
\subfloat[Stackoverflow]{\includegraphics[width=0.45\linewidth]{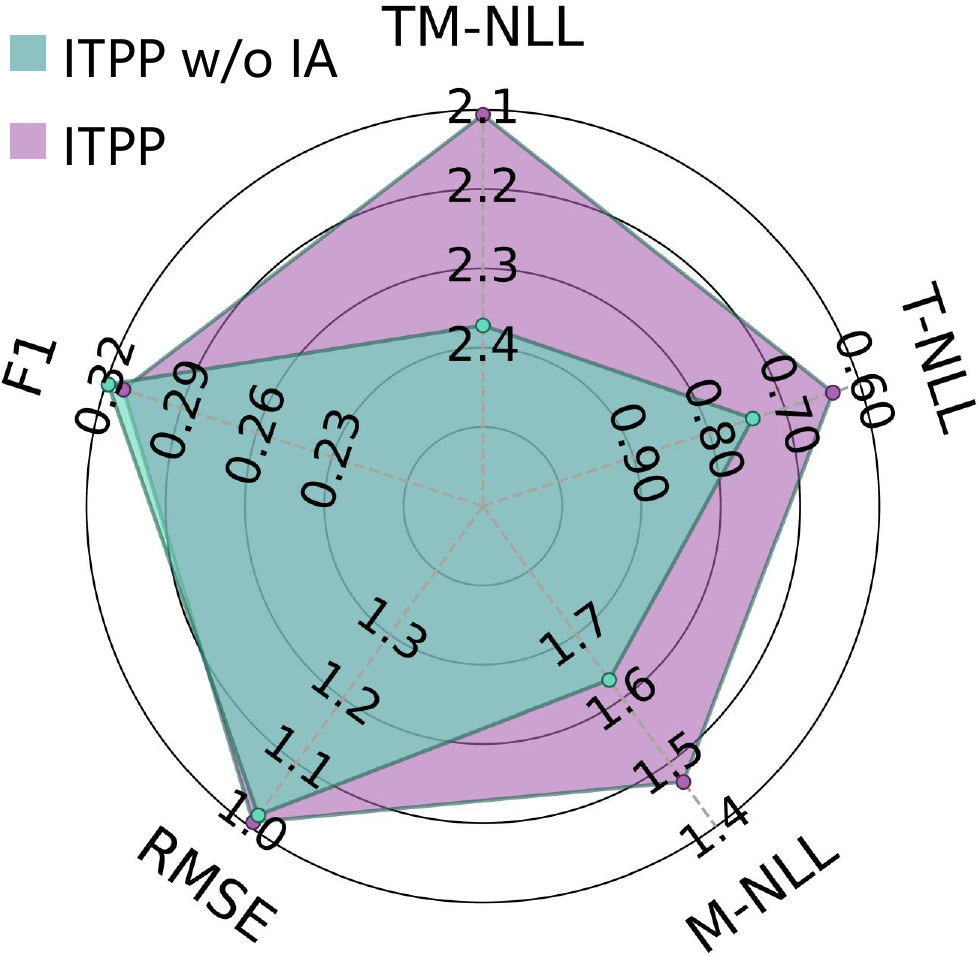}}
\hfil
\subfloat[Earthquake]{\includegraphics[width=0.45\linewidth]{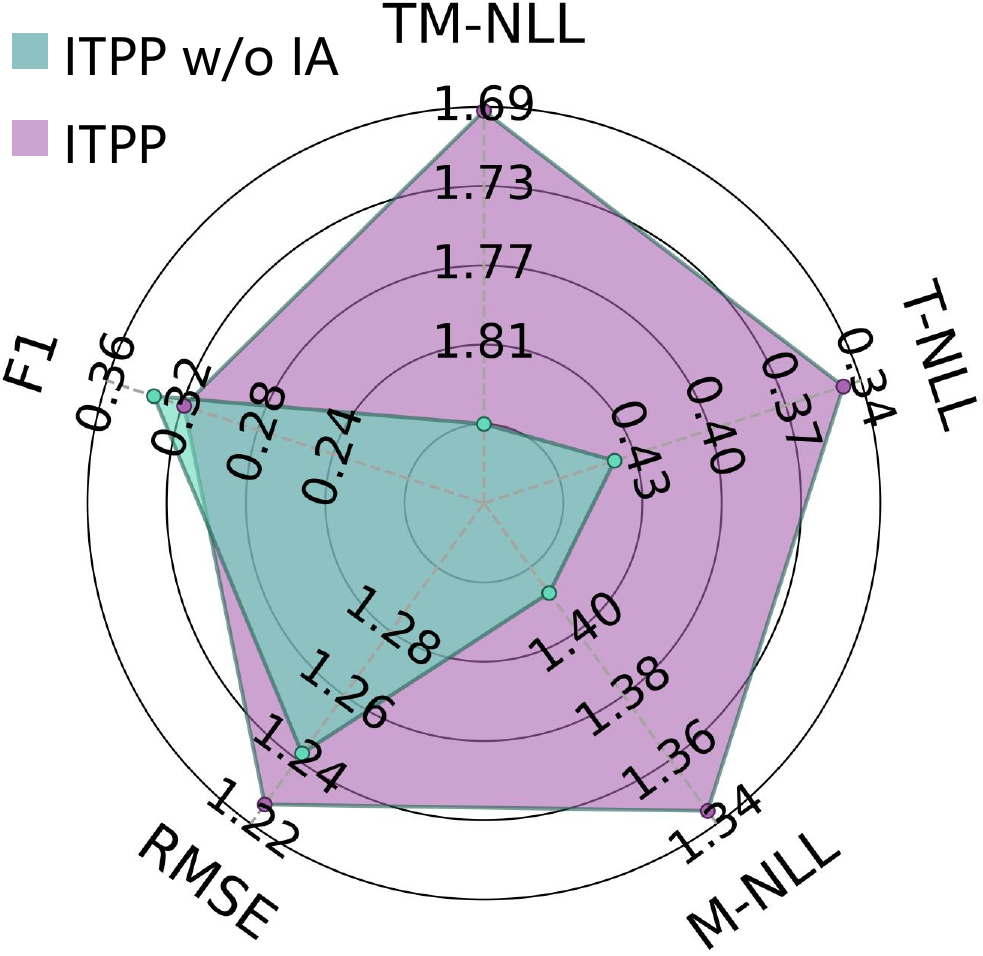}}
\caption{Ablation study results of inverted self-attention.}
\label{fig:ablation-ia}

\end{figure}

Next, we evaluate the contribution of the inverted self-attention module, which is designed to capture correlations between different event channels. To achieve this, we compare the full ITPP model against a variant that lacks this module, with the results presented in Fig. \ref{fig:ablation-ia}.
We find that removing the inverted self-attention module causes a substantial drop in model performance, which suggests that different event types in the datasets are highly correlated and that failing to capture these dependencies undermines the model's predictive power.

\subsection{Overfitting}

\begin{figure}[tb]

\centering
\subfloat[Training loss]{\includegraphics[width=0.47\linewidth]{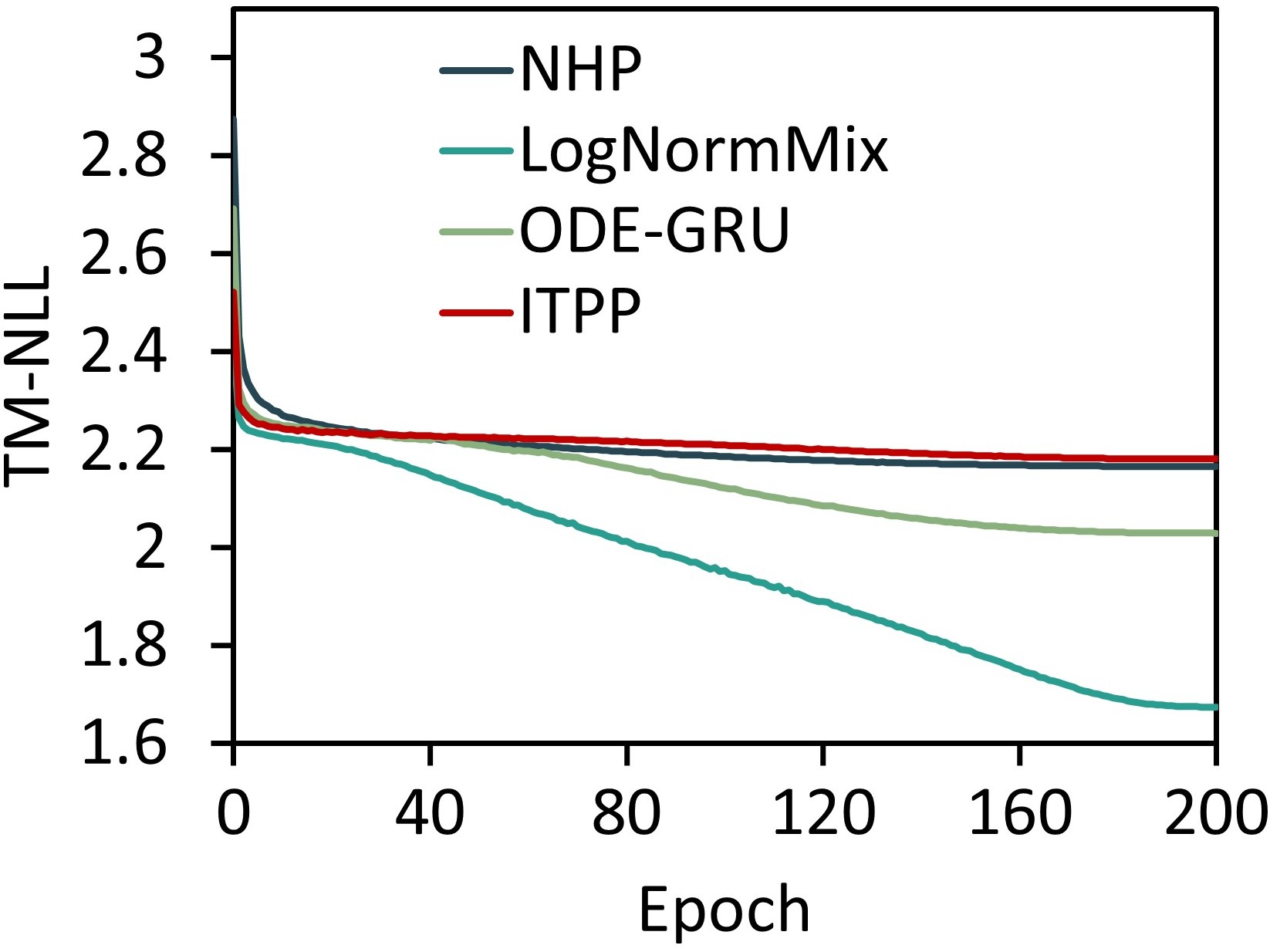}}
\hfil
\subfloat[Testing loss]{\includegraphics[width=0.47\linewidth]{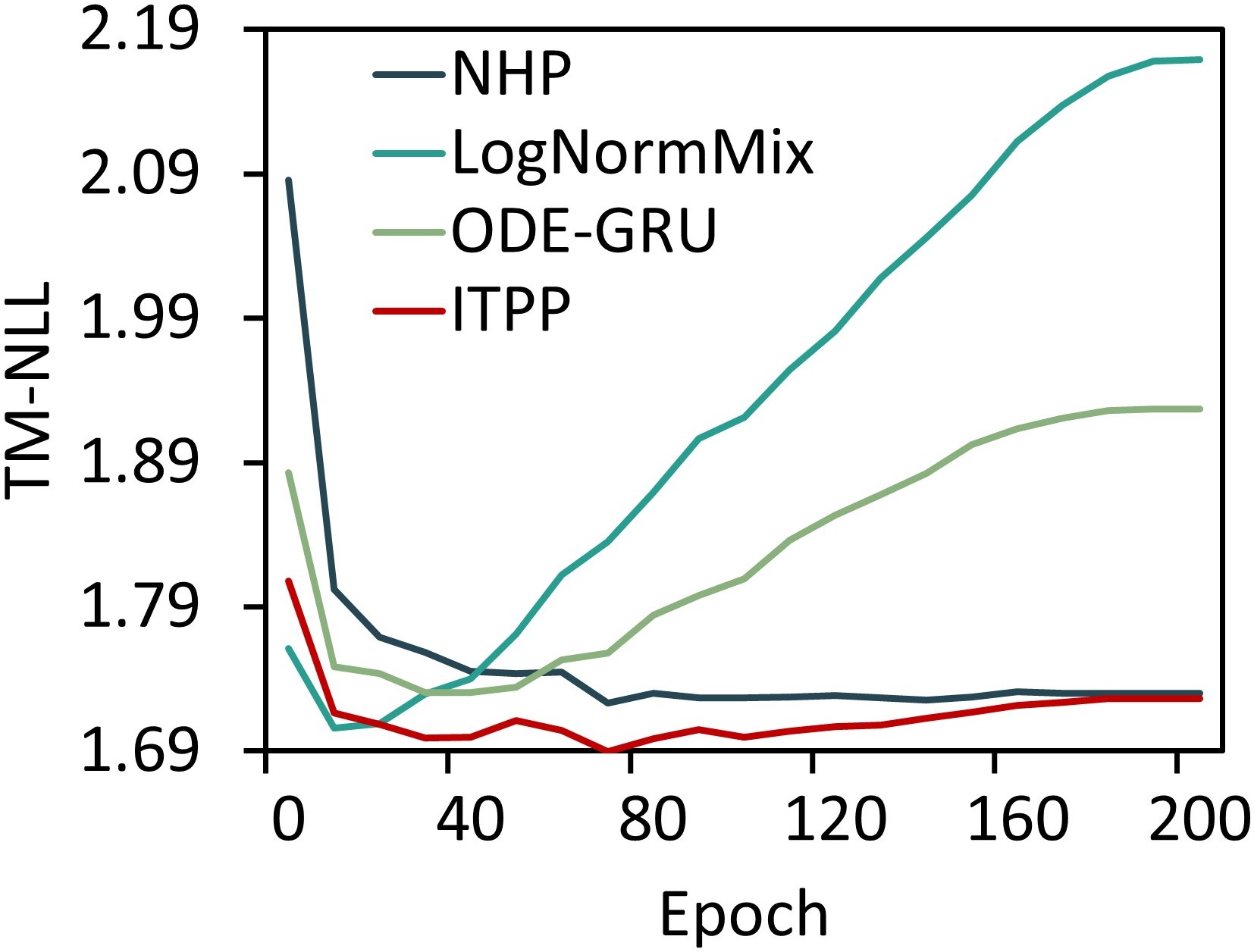}}
\caption{The changes of training and testing loss as the training progresses.}
\label{fig:overfitting}

\end{figure}

Overfitting is a significant challenge when training MTPP models, particularly on smaller datasets. 
Fig. \ref{fig:overfitting} visualizes the training and testing loss curves for ITPP and three baseline models on the Earthquake dataset.
The figure reveals severe overfitting in LogNormMix and ODE-GRU, where their training losses decrease steadily while the testing loss starts to surge after some point. This behavior complicates training and compromises the models' robustness for real-world applications. The proposed ITPP, however, by enforcing information disentanglement and explicit correlation capturing, demonstrates stronger resistance to overfitting.

\section{Conclusion}
This paper introduces a novel channel-independent framework for MTPP modeling, addressing the limitations of conventional channel-mixing approaches. By leveraging a multi-channel, ODE-based architecture, our model effectively disentangles and simulates the distinct temporal dynamics of each event type. A type-aware inverted self-attention mechanism is proposed to explicitly model inter-type dependencies, enhancing the model's expressiveness. Extensive empirical evaluations demonstrate that our approach improves robustness and generalization and sets a new benchmark in MTPP modeling performance.

\section{Acknowledgments}
This research work was partly supported by National Natural Science Foundation of China under grant No. 62376055

\bibliography{bibitems}

\clearpage

\appendix
\section*{Appendix}

\section{A. Implementation details}
Our implementation is built upon the JAX 0.5.3 framework \cite{jax} with Python 3.11.5 and CUDA 11.6, utilizing Equinox 0.12.2 \cite{equinox} for model creation. Differential equations are solved and optimized using the Tsit5 solver \cite{tsit} provided by the Diffrax 0.7.0 package \cite{diffrax}. We optimized the models using the AdamW optimizer provided by Optax 0.2.4 \cite{optax} with an initial learning rate of 0.001 on a workstation equipped with an Intel Xeon Gold 5218 CPU and NVIDIA GeForce RTX 4090 GPUs. All experimental results are averaged across 5 runs.

\vspace{-0.5em}

\section{B. Datasets}
This section details the datasets used for evaluation, including both real-world and synthetic data. For the real-world datasets, we describe their original sources and context. For the synthetic datasets, we outline their generation process.

\subsection{B.1 Real-world datasets}
\paragraph{StackOverflow \cite{stackoverflow}} This dataset captures all badges granted to users over a two-year period on a knowledge-sharing website. The data is structured as sequences of awards for each user, with 22 unique badge types in total. A subset of the original dataset is used in our experiments, with a training set of 1401 sequences, a validation set of 401 sequences and a testing set of 401 sequences.

\paragraph{MIMIC \cite{mimic}} This dataset consists of a seven-year collection of de-identified patient records from an Intensive Care Unit (ICU). The data is organized into sequences of clinical visits for each patient, with every visit containing a timestamp and its corresponding disease diagnosis. There is a total of 75 types of disease in the subset used in the experiments, which is divided into a training set of 527 sequences, a validation set of 58 sequences and a testing set of 65 sequences.

\paragraph{Taobao \cite{hypro}} This dataset, first released for the 2018 Tianchi Big Data Competition, consists of timestamped user activities (e.g., Browse, purchasing) on the Taobao platform from November 25 to December 3, 2017. We treat product categories as event types, resulting in a total of 17 distinct types. The dataset is partitioned into disjoint training, validation, and testing sets of 1300, 200, and 500 sequences.

\paragraph{Earthquake \cite{easytpp}} Sourced from the Earthquake Hazards Program of United States Geological Survey (USGS), this dataset contains a record of earthquake events in the Conterminous U.S. between 1996 and 2023. The 7 distinct event types correspond to different earthquake magnitudes. The data is partitioned into a training set of 3000 sequences, a validation set of 400 sequences and a testing set of 900 sequences.

\subsection{B.2 Synthetic datasets}
\paragraph{Poisson} This dataset was synthetically generated using Ogata's thinning algorithm \cite{thinning} to simulate a multivariate homogeneous Poisson process over the time interval [0,10]. The process consists of 3 event types with constant intensities set to $ \mu_1=5 $, $ \mu_2=1 $ and $ \mu_3=2 $. We created 500 sequences, which were then split into training, validation, and test sets in a 3:1:1 proportion.

\paragraph{Hawkes} We generated this dataset using a multivariate self-exciting Hawkes process \cite{hawkes} whose intensity function is defind as follows:
\begin{equation}
    \lambda_k(t)=\mu_k+\sum_{i;t_i<t}\alpha_{i,k}\cdot\exp(-\phi_{i,k}(t-t_i))
\end{equation}
where $ \mu_k $ is the constant base rate of event type $ k $, while $ \alpha_{i,k} $ and $ \phi_{i,k} $ are the peak exciting rate and decaying rate, respectively, of event type $ i $ upon type $ k $. Specifically, we refined the number of event types to be 3, i.e., $ k\in \{1, 2, 3\} $, and set the parameters to be:
\begin{equation}
    \boldsymbol{\mu}=
    \begin{bmatrix}
        0.5\\
        0.4\\
        0.3
    \end{bmatrix},
    \boldsymbol{\alpha}=
    \begin{bmatrix}
        1 & 0.5 & 0.5\\
        0.5 & 1 & 0.5\\
        0.5 & 0.5 & 1
    \end{bmatrix},
    \boldsymbol{\phi}=
    \begin{bmatrix}
        2 & 4 & 4\\
        4 & 2 & 4\\
        4 & 4 & 2
    \end{bmatrix}
\end{equation}
The sequences were also generated using the thinning algorithm \cite{thinning} within the time range of [0,10]. A total of 500 sequences were generated and partitioned into training, validationa and testing sets in a proportion of 3:1:1.

\vspace{-0.5em}

\section{C. Baselines}
In this section, we give a brief introduction of all the baseline models used for comparison in our experiments.

\subsection{C.1 RNN-based models}

\paragraph{RMTPP \cite{rmtpp}} This is one of the earliest neural MTPP works. They propose to to encode asynchronous event sequences with a modified RNN network. The intensity function is elaborately designed so that log-likelihood of event sequences can be computed in closed form, facilitating efficient training.

\paragraph{NHP \cite{nhp}} They extend RMTPP to a more sophisticated continuous-time LSTM framework, which mimics the formulation of self-exciting Hawkes processes \cite{hawkes}. While achieving better flexibility in modelling, the training efficiency is compromised due to a numerical estimation of the log-likelihood.

\paragraph{LogNormMix \cite{lognormmix}} This work focus on the decoder side of the MTPP framework. Instead of modeling the intensity function, they propose to fit the target distribution with a mixture of Log-Normal distributions. This design avoids the computation cost of numerical estimation of log-likelihood, and achieves great performance boost in probabilistic fitting.

\subsection{C.2 Self-attentive models}

\paragraph{SAHP \cite{SAHP}} Inspired by the success of Transformers \cite{transformer} in the area of NLP, this work devise a self-attentive encoder for MTPP modelling. The core of this model is a time-shifted positional embedding, that encodes irregular time intervals of MTPP event sequences.

\paragraph{THP \cite{THP}} This is a parallel work that introduces the self-attention mechanism to MTPP modelling. It differs slightly from SAHP in terms of the design of temporal encoding and intensity formulation.

\paragraph{AttNHP \cite{attnhp}} This work derives from the idea of NHP \cite{nhp} and develops a more sophisticated hierarchical event embedding mechanism based on a multi-layer self-attention network to incorporate temporal and mark information. 

\begin{algorithm}[t]
\caption{Training of ITPP}
\label{alg:training}
\textbf{Input}: Event sequences $ \{\mathcal{S}_j\}_{j=1}^N $, where $ \mathcal{S}_j=\{(t_i^j, m_i^j)\}_{i=1}^L $\\
\textbf{Parameters}: $ \boldsymbol{\theta}=\{\boldsymbol{\theta}_f,\boldsymbol{\theta}_g,\boldsymbol{\theta}_c,\boldsymbol{\theta}_r\} $
\begin{algorithmic}[1] 
\WHILE{NOT CONVERGED}
\FOR{$j\gets 1$ to $ N $}
\STATE $\boldsymbol{Z}\gets$INIT\_STATE; $ \Lambda\gets0$; $t\gets0 $; $\eta\gets0$.
\FOR{$i \gets 1$ to $L$}
\STATE $[\boldsymbol{Z}$, $\Lambda]\gets$ODESolve$([\boldsymbol{f}_{\boldsymbol{\theta}_f},\boldsymbol{r}_{\boldsymbol{\theta}_r}],[\boldsymbol{Z},\Lambda],t,t_i^j)$.
\STATE $\eta\gets\eta+\log(\boldsymbol{r}_{\boldsymbol{\theta}_r}(\boldsymbol{Z}[m_i^j]))$.
\STATE $\boldsymbol{Z}[m_i^j]\gets \boldsymbol{g}_{\boldsymbol{\theta}_g}(\boldsymbol{Z}[m_i^j])$.
\STATE $ t\gets t_i^j $.
\ENDFOR
\STATE $ \mathcal{L}=\Lambda-\eta $.
\STATE Back-propagate with gradien $ \nabla_{\boldsymbol{\theta}}\mathcal{L} $.
\ENDFOR
\ENDWHILE
\end{algorithmic}
\end{algorithm}

\subsection{C.3 CNN-based model}
\paragraph{CTPP \cite{ctpp}} This work derives from LogNormMix \cite{lognormmix}, and introduces a novel continuous-time convolutional neural network to better capture local event contexts. This framework is flexible and scalable and can focus on the local context of different horizons.

\subsection{C.4 DE-based models}
\paragraph{Neural ODE \cite{neuralode}} This work first proposed using neural networks to learn the complex dynamics of continuous states. While not originally intended for MTPP modeling, we adapt it as a baseline for comparison. Our adaptation involves adding a decoder to map the continuous latent state to event intensities. This provides a direct contrast to our model, highlighting the benefits of our specific design choices.

\paragraph{ODE-GRU \cite{gruode}} This work derives from Neural ODE by using a GRU-based design to construct the drift network. Similarly, this work was not intended for MTPP. Some adaptation was made to fit the intensity modelling scenario.

\paragraph{NJDTPP \cite{njdtpp}} This work presents an SDE-based MTPP model with two main contributions. First, it adds a stochastic component to the governing differential equation to account for uncertainty. Second, its dynamics operate directly on the logarithm of the intensity values, bypassing the need for an intermediate latent state. 

\vspace{-0.3em}

\section{D. Training algorithm}

Algorithm \ref{alg:training} shows the training process of ITPP. $ \boldsymbol{Z} $ is a real-value matrix of shape $ K\times d $, each row of which represents the dynamic state of an independent channel. $ \eta $ is a cumulative scalar variable used to calculate the first term of the log-likelihood formula (Eq. 11), while $ \Lambda $ is a scalar variable that keeps track of the integral term. Given an event sequence, the algorithm simulates the state evolution from t=0 to the last event, and computes the negative log-likelihood of the whole sequence as the training loss $ \mathcal{L} $, which is then used for back-propagation. The ODESolve function is a black-box ODE solver that simulates the state extrapolation (Eq. 3) and computes the intensity integral in the meantime. 

\begin{table}[t]
  \setlength{\tabcolsep}{0pt}
  \begin{tabular*}{\columnwidth}{@{\extracolsep{\fill}}l *{3}{c}}
    \toprule
    Methods & TM-NLL & T-NLL & M-NLL \\
    \midrule
    RMTPP  & 0.411 & -0.668 & 1.078  \\
    NHP    & 0.330 & -0.746 & 1.076 \\
    LogNormMix  & 0.329 & -0.747 & 1.076 \\
    THP  & 0.388 & -0.704 & 1.091 \\
    SAHP  & 0.354 & -0.731 & 1.084\\
    NeuralODE  & 0.328 & \textit{-0.750} & 1.078 \\
    ODE-GRU & \underline{0.323} & \underline{-0.752} & 1.075 \\
    CTPP  & 0.331 & -0.744 & \underline{1.074}\\
    AttNHP & \textit{0.327} & -0.748 & \underline{1.074} \\
    NJDTPP & 0.335 & -0.744 & 1.079 \\
    \midrule
    \textbf{ITPP} & \textbf{0.316} & \textbf{-0.755} & \textbf{1.071} \\
    \bottomrule
  \end{tabular*}
  \caption{Probabilistic evaluation results on Hawkes datasets}
  \label{tab:syn-nll}
\end{table}

\begin{table}[t]
  \setlength{\tabcolsep}{0pt}
  \begin{tabular*}{\columnwidth}{@{\extracolsep{\fill}}l *{2}{c}}
    \toprule
    Methods & RMSE & F1  \\
    \midrule
    RMTPP  & 0.320 & \textit{0.416}   \\
    NHP    & \underline{0.316} & 0.411 \\
    SAHP  & 0.317 & 0.399\\
    NeuralODE  & 0.317 & 0.407 \\
    ODE-GRU & \underline{0.316} & \underline{0.421} \\
    NJDTPP & 0.318 & 0.389 \\
    \midrule
    \textbf{ITPP} & \textbf{0.315} & \textbf{0.423} \\
    \bottomrule
  \end{tabular*}
  \caption{Prediction evaluation results on Hawkes datasets}
  \label{tab:syn-pred}
  \vspace{-1em}
\end{table}

\vspace{-0.3em}

\section{E. Supplementary results on synthetic datasets}
Table \ref{tab:syn-nll} and Table \ref{tab:syn-pred} shows the probabilistic and prediction evaluation results on Hawkes dataset, respectively. ITPP outperform all baseline models on all five metrics. The channel-mixing counterpart ODE-GRU comes second, falling a little behind ITPP, probably caused by the entanglement of information. These results are generally consistent with those of real-world datasets given in the paper.






\end{document}